\documentclass{article}

\usepackage{microtype}
\usepackage{graphicx}
\usepackage{subcaption}
\usepackage{booktabs} % for professional tables
\usepackage{hyperref}
\usepackage[preprint]{icml2026}

\usepackage{algorithm}
\usepackage{amsmath}
\usepackage{amssymb}
\usepackage{mathtools}
\usepackage{amsthm}
\usepackage{soul}
\usepackage{color}
\usepackage{xcolor}
\usepackage{amsmath}
\usepackage{amsfonts}
\usepackage{multirow}
\usepackage{xspace}
\usepackage{booktabs}
\usepackage{enumitem}
\usepackage{nameref}
\usepackage{array}
\usepackage{nicematrix} 
\usepackage{appendix}
\usepackage{titletoc}
\usepackage{pifont}

\newcommand{\method}{\texttt{RepSPD}\xspace }

\usepackage[capitalize,noabbrev]{cleveref}

\theoremstyle{plain}

\theoremstyle{definition}

\theoremstyle{remark}

\usepackage[textsize=tiny]{todonotes}

\icmltitlerunning{RepSPD: Enhancing SPD Manifold Representation in EEGs via Dynamic Graphs}

\begin{document}

\twocolumn[
  \icmltitle{RepSPD: Enhancing SPD Manifold Representation in EEGs \\via Dynamic Graphs}




  \begin{icmlauthorlist}
    \icmlauthor{Haohui Jia}{yyy}
    \icmlauthor{Zheng Chen}{comp}
    \icmlauthor{Lingwei Zhu}{sch}
    \icmlauthor{Xu Cao}{lab}
    \icmlauthor{Yasuko Matsubara}{comp}
    \icmlauthor{Takashi Matsubara}{yyy}
    \icmlauthor{Yasushi Sakurai}{comp}
    %\icmlauthor{Firstname8 Lastname8}{sch}
    %\icmlauthor{Firstname8 Lastname8}{yyy,comp}
  \end{icmlauthorlist}

  \icmlaffiliation{yyy}{Information Science and Techinology, Hokkaido University, Japan}
  \icmlaffiliation{comp}{SANKEN, The University of Osaka, Japan}
  \icmlaffiliation{sch}{Great Bay University, China}
  \icmlaffiliation{lab}{Department of Computer Science, University of Illinois Urbana-Champaign, USA}
  \icmlcorrespondingauthor{}{chenz@sanken.osaka-u.ac.jp}
  % \icmlcorrespondingauthor{Firstname1 Lastname1}{first1.last1@www.uk}

  \icmlkeywords{Machine Learning, ICML}

  \vskip 0.3in
]



%\printAffiliationsAndNotice{}  % no special notice (required even if empty)
\printAffiliationsAndNotice{}

\begin{abstract}
Decoding brain activity from electroencephalography (EEG) is crucial for neuroscience and clinical applications. Among recent advances in deep learning for EEG, geometric learning stands out as its theoretical underpinnings on symmetric positive definite (SPD) allows revealing structural connectivity analysis in a physics-grounded manner.
However, current SPD-based methods focus predominantly on statistical aggregation of EEGs, with frequency-specific synchronization and local topological structures of brain regions neglected.
Given this, we propose \method, a novel geometric deep learning (GDL)-based model.
\method implements a cross-attention mechanism on the Riemannian manifold to modulate the geometric attributes of SPD with graph-derived functional connectivity features. On top of this, we introduce a global bidirectional alignment strategy to reshape tangent-space embeddings, mitigating geometric distortions caused by curvature and thereby enhancing geometric consistency.
Extensive experiments demonstrate that our proposed framework significantly outperforms existing EEG representation methods, exhibiting superior robustness and generalization capabilities.
\end{abstract}

\section{Introduction}
\label{sec:introduction}
% A brain-computer interface (BCI) is designed to extract meaningful patterns from electroencephalography (EEG) signals, enabling a wide range of applications, including rehabilitation and communication \cite{mcfarland2011brain}. 
% However, current BCIs face significant challenges, such as low signal-to-noise ratio, limited specificity, and non-stationary signal characteristics \cite{fairclough2020grand}.
% To address these challenges, modern EEG-based BCIs typically utilize machine learning to analyze brain signals; consequently, the effectiveness of these interfaces is heavily influenced by the reliability and accuracy of EEG decoders \cite{lotte2018review}.

% Currently, the state-of-the-art in EEG-BCI decoding is typically achieved through two main, yet relatively independent, methodological approaches: deep learning \cite{schirrmeister2017deep, lawhern2018eegnet} and Riemannian-geometry-based (RBD) decoders \cite{barachant2011multiclass, zanini2017transfer}.
% Deep learning relies on multi-layer neural network architectures that are trained via backpropagation, enabling these models to uncover complex, hierarchical patterns within raw data automatically. 
% This approach has driven major advances across various domains, including EEG decoding \cite{lecun2015deep}.

Learning functional connectivity in the brain is essential for understanding how different brain areas coordinate their activity to perform various cognitive functions and physiological tasks.
This understanding is typically facilitated by statistical methods, often built upon neuroimaging technologies such as electroencephalography (EEG) \cite{PNAS2014,NatMecinecom}.

Among a plethora of methods, the Riemannian manifolds defined by the Symmetric Positive Definite (SPD) matrices have shown both empirical promises and theoretical rigor in representing brain connectivity patterns \cite{NEURIPS2022_28ef7ee7,ju2024deep}.
SPD matrices model the second-order spatial statistics (i.e., covariance) of EEGs to reveal how activities from different brain regions co-vary, thereby reflecting functional connectivity.
Moreover, these matrices have also been conceptualized as points on SPD manifolds when equipped with a Riemannian metric to map the intrinsic geometry of brain connectivity patterns \cite{pennec2020manifold}.
% such as affine-invariant \cite{pennec2020manifold} or log-Euclidean metric \cite{huang2017riemannian}.
In other words, SPD matrices can be regarded as EEG-based representations of functional connectivity, where the location of an SPD point on the manifold reflects a specific brain functional state (i.e., \a class indicator of the representation).

\begin{figure}[!t]
\centering
\includegraphics[width = 0.99\linewidth]{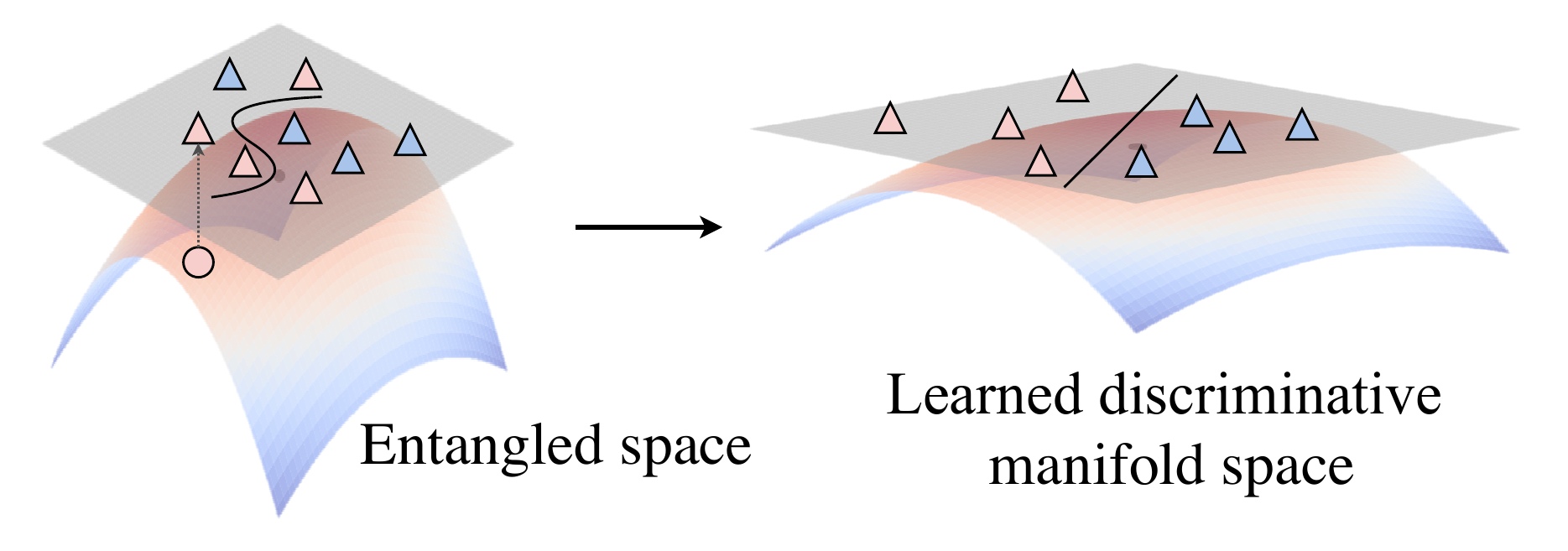}
\caption{Concept of SPD point projections from the manifold to tangent space. Our goal is to modulate SPD representations to achieve effective tangent embeddings.}
\vspace{-0.5 cm}
\label{fig: intro}
\end{figure}
% Such powerful methods have been used in various application, including motor imagery, disease detection, and xxxx.
% However, classic SPD-based works remain constrained by modeling flexibility and representation capacity. The handcrafted statistical construction of SPD matrices fails to capture the nonlinear dynamics of EEG signals from trials and subjects.

However, the expressiveness of SPD manifold representations remains constrained, since the covariance estimation is rooted in linear statistics and fails to model the nonlinear and task-specific nature of EEGs.
To tackle this issue, researchers have proposed geometry-aware deep networks to learn the SPD matrices in a data-driven manner yet preserve the Riemannian structure.
\cite{Suh_Kim_2021,ju2022graph,li2024spdim}
By projecting the manifold representations back to the Euclidean space, these methods can naturally perform downstream tasks like classification as usual models.
However, existing methods focus mostly on mathematical tractability, ignoring explicit modeling of brain dynamics such as functional connectivity.
On one hand, SPD treats all channels equally without any weighting or intra-channel geometry modeling;
on the other hand, these methods exclusively deal with energy-attributed connectivity with covariance aggregation by assuming temporal stationarity within each EEG epoch.
As a result, their representations ignore local dynamics and  time-course of brain connectivity.
% First, although models like BiMap \cite{huang2017riemannian} enable nonlinear learning, the SPD construction treats all channels equally, without incorporating any form of channel weighting or inference of intra-channel geometric structure.
% While few works segment EEG epochs into fine-grained short snapshots to model structural varying from sequential snapshots \cite{pan2022matt}, they still fail to represent temporal dependencies or inter-structure dynamics.
% Therefore, the objective of this paper is to explicitly model dynamic brain functional connectivity by learning structure-aware SPD representations.
Given this, the goal of this paper is to explicitly model dynamic brain functional connectivity by learning structure-aware representations that naturally fit into the SPD Riemannian framework.

We propose \method, a deep learning framework that integrates dynamic graph neural networks (GNN) into the Riemannian SPD manifold learning workflow.
Specifically, we propose a dual-SPD neural network that simultaneously learns two complementary connectivity representations from EEGs:
a manifold representation from covariance estimation; and a graph-based representation from the time-then-graph paradigm \cite{kotoge2025evobrain}.
The graph representation models dynamic dependencies and embeds functional connectivity to the manifold in an adaptive manner.
However, a challenge arises with this integration as the graph representation is learned in the Euclidean space, not satisfying the constraints of the Riemannian manifold.
One of our contributions consists in a novel Dynamic Manifold Attention network module that maps graph-derived features into a Riemannian geometry compatible form, thereby enabling representation fusion  in a geometry-preserving way.
This is achieved by a novel geometry alignment loss that mitigates the geometric inconsistency between the two latent spaces.
This loss further addresses the distortion caused by the positive curvature of the manifold, where samples may become overly dispersed in the tangent space.
% Our ablation study demonstrates that this loss design  facilitates more discriminative decision boundaries for downstream tasks.

Overall, the contributions of this work lie in:
\ding{172} \textbf{Conceptually}, we combine the best of both worlds from the Riemannian manifolds and GNN, integrating dynamic modeling of brain functional connectivity into the mathematically rigorous Riemannian geometry, bridging statistical modeling with graph-based representations.
\ding{173} \textbf{Methodologically}, we propose an end-to-end framework to integrate the Euclidean dynamic graph representations into the Riemannian manifolds while ensuring the fused representation respects positive definiteness. This integration technique is of independent interest to researchers pursuing non-Riemannian integration to the SPD frameworks.
\ding{174} \textbf{Practically}, we conduct extensive experiments on the BCI and clinical seizure detection datasets to verify the effectiveness of \method.

Our \method outperforms all baselines on the TUSZ dataset, achieving a $3.3\%$ improvement. On the BCI dataset, \method consistently attains the best performance, with up to a $2.23\%$ accuracy gain.
Importantly, these improvements are not merely numerical; they arise from (i) leveraging explicit dynamic topology to complement SPD-based statistics, and (ii) enforcing geometry-consistent fusion and cross-view alignment, which stabilizes discriminability in the tangent space.
\section{Related Works}
\label{sec:related_works}

\noindent\textbf{SPD Matrices and Networks in EEGs. }
SPD matrices employ concepts from Riemannian geometry to leverage the inherent geometrical properties of the covariance matrices derived from EEGs \cite{barachant2010riemannian,zhang2020manifold,DeepRiemannian2025}.
In recent years, several works aim to combine SPD with deep neural networks and propose various Deep SPD Networks \cite{ju2025spdlearningcovariancebasedneuroimaging}.
Some of these methods have been demonstrated on facial and image recognition datasets \cite{Dong_Jia_Zhang_Pei_Wu_2017,Acharya_2018_CVPR_Workshops,Liu_Li_Wu_Ji_2019,wu2022fusion}, and also some have been applied to EEG tasks \cite{Liu_Li_Wu_Ji_2019,Suh_Kim_2021,ju2022tensor,kobler2022spd,zhang2023spatio,ju2024deep}.
One milestone method is SPDNet \cite{huang2017riemannian} that mimics certain functions of convolutional networks but operates entirely on SPD matrices.
Several follow-up models based on SPDNet have since been proposed to address various EEG tasks, including challenges such as label shift \cite{li2024spdim}, domain adaptation \cite{kobler2022spd}, and subject-transfer learning \cite{9204368}.

\noindent\textbf{Graph Neural Networks in EEGs. }
Graph neural networks (GNNs) have emerged as powerful methods for effectively capturing spatially dynamic dependencies in the analysis of brain functional connectivity \cite{9630194,GNN_AAAI23,kan2023r,pmlr-v209-tang23a}.
Specifically, 
ST-GCN formulates the connectivity of spatio-temporal graphs to capture non-stationary changes in neural dynamics \cite{gadgil2020spatio}.
Siyu et al. \cite{GNN_ICLR22} introduced the DCRNN approach for graph modeling, setting a new standard for state-of-the-art (SOTA) performance in seizure detection and classification.
% Building on this, GRAPHS4MER \cite{pmlr-v209-tang23a} enhanced the graph structure and integrated it with the MAMBA framework to improve long-term modeling capabilities.
Additionally, the recent AMAG framework \cite{li2024amag} was proposed to effectively capture causal relationships in brain functional connectivity, demonstrating greater efficiency in modeling temporal dynamics.
While existing methods focus primarily on feature extraction and representation learning, our work takes a different perspective: we leverage dynamic GNNs to enhance the representation quality of the SPD manifold.

\begin{figure*}[t]
\centering
\includegraphics[width = 0.99\textwidth]{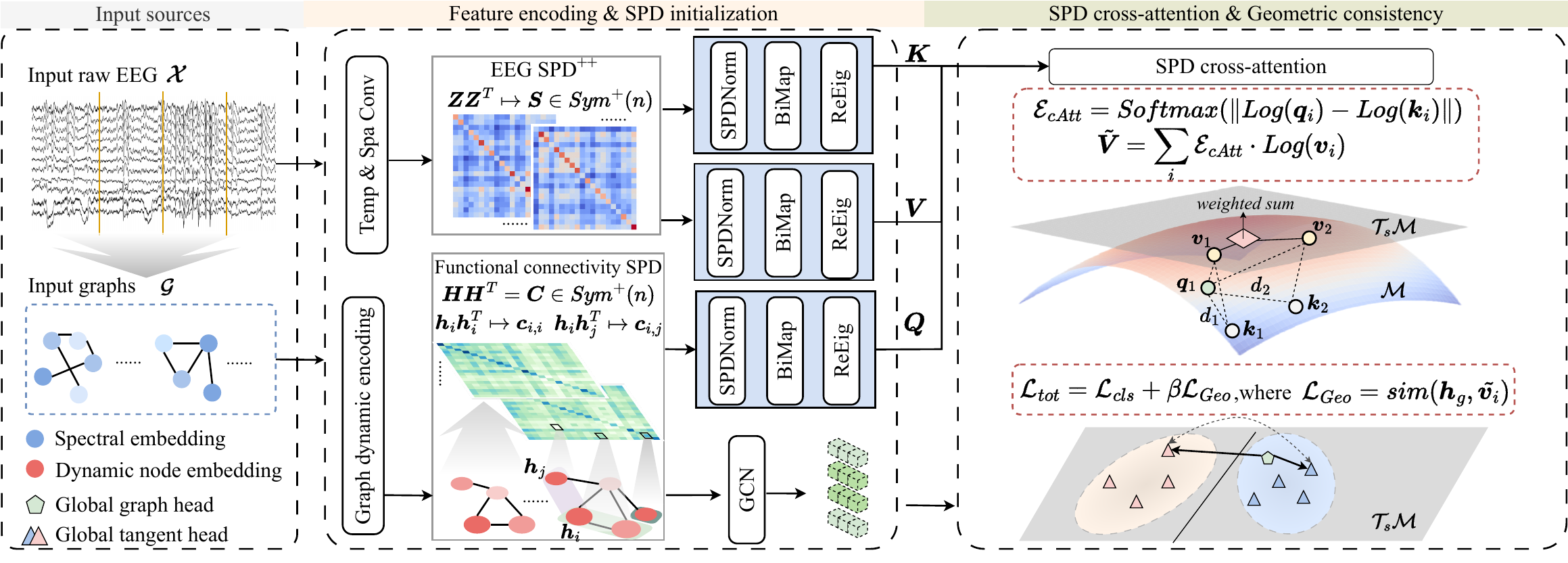}
\caption{Overview of the proposed \method.
Raw EEG signals and graph-constructed functional connectivity are separately encoded into SPD matrices. These SPD representations are processed via SPDNet to generate query, key, and value matrices for modulating the SPD with cross-attention. Our dynamic graph guided modulation operates on the Riemannian SPD manifold, computing attention weights based on Log-Euclidean distance and modulating the EEG representation with graph structures. A structure-aware loss function, combining task loss and a geometry-level alignment loss $\mathcal{L}_{GeoTop}$, ensures functional-structural consistency and robust discriminative representations in tangent space.}
\label{fig: proposal}
\end{figure*}
\section{Preliminary and Problem Formulation}
\label{sec:problem}

We consider modeling the given EEG signals as manifolds.  
Specifically, we first formulate the problem and define the learning objective based on the Riemannian metrics on $\mathbb{S}_{++}^{N}$, followed
by an analysis of the fundamental concepts underlying the SPD manifold.

\noindent\textbf{Definition 1 (EEG Data and Functional Connectivity).}  
We define an EEG recording can be segmented a sequence of $K$ consecutive samples as $\mathcal{X} = [\boldsymbol{X}_{1},\boldsymbol{X}_{2},\dots,\boldsymbol{X}_{K}]$, each $\boldsymbol{X}_{k} \in \mathbb{R}^{N \times L}$ contains $N$ channels and $L$ time points.  
To capture brain dynamics, we apply a sliding window to segment $\boldsymbol{X}_k$ into a sequence of EEG refined epochs, e.g., 1-second segments \cite{SODorAAAI2025}.  
EEG functional connectivity refers to the statistical dependency between spatially distributed brain regions (channels).  
Formally, the functional connectivity between $v_i$ and $v_j$ at epoch $t$ is quantified by a similarity function $\phi(\boldsymbol{h}_{i,t}, \boldsymbol{h}_{j,t})$, such as cosine similarity or Riemannian distance between SPD matrices.

\noindent\textbf{Definition 2 (EEG Graph).}   
We define an EEG graph $\mathcal{G} = \{ \mathcal{V}, \mathcal{A}, \boldsymbol{X} \}$, where $\mathcal{V} = \{ \boldsymbol{v}_1, \dots, \boldsymbol{v}_N \}$ denotes $N$ channels (nodes), and $\mathcal{A} \in \mathbb{R}^{N \times N \times T}$ denotes the temporal adjacency tensor.  
Each $a_{i,j,t}$ reflects the functional similarity between channels $\boldsymbol{v}_i$ and $\boldsymbol{v}_j$ at the $t$-th epoch, based on their representations, e.g., spectrum.  
The node embedding $\boldsymbol{h}_{i,t} \in \mathbb{R}^m$ represents the feature vector of channel $\boldsymbol{v}_i$ at epoch $t$.\\
\textit{- Graph connectivity.} The resulting pairwise connectivity values form the epoch-wise adjacency matrix $\mathcal{A}_t \in \mathbb{R}^{N \times N}$, capturing the dynamic interactions among channels.

\noindent\textbf{Definition 3 (SPD Riemannian Manifold).}  
Let $\mathbb{S}_{++}^N$ denote the set of $N \times N$ SPD matrices.  
This set forms a Riemannian manifold $(\mathcal{M}, g)$, where $\mathcal{M} = \mathbb{S}_{++}^N$ and $g$ is a Riemannian metric, such as the affine-invariant or Log-Euclidean metric.

For any point $\boldsymbol{C} \in \mathcal{M}$, the tangent space at $\boldsymbol{C}$ is denoted as $T_{\boldsymbol{C}}\mathcal{M}$, representing a local Euclidean approximation of the manifold.  
The logarithmic map and exponential map define the projection between the manifold and its tangent space, allowing SPD representations to be projected into Euclidean features that can be utilized by downstream models such as classifiers.
To ensure geometric consistency between the manifold $\mathcal{M}$ and the tangent space $T_{\boldsymbol{C}}\mathcal{M}$, the geodesic distance between two SPD matrices $\boldsymbol{C}_1, \boldsymbol{C}_2 \in \mathbb{S}_{++}^N$ under the Log-Euclidean metric is defined as:
\[
d_{\text{LE}}(\boldsymbol{C}_1, \boldsymbol{C}_2) = \left\| \log(\boldsymbol{C}_1) - \log(\boldsymbol{C}_2) \right\|_F,
\]
where $\log(\cdot)$ denotes the matrix logarithm,
and $\log(\boldsymbol{C})$ maps the SPD from the manifold to a Euclidean space while preserving its geometric structure, e.g., $d_{\text{LE}}$.\\
\textbf{Remarks. }In this work, this metric is not only used to ensure geometric consistency on the SPD manifold, but also serves as a critical operator to bridge SPD-based and graph representations for feature fusion.\\
\textit{- Manifold connectivity.} 
In the context of EEG, a sequence $\{ \boldsymbol{C}_t \}_{t=1}^{T}$ of covariance matrices computed from consecutive EEG epochs forms a trajectory on manifold $\mathcal{M}$.
Each $\boldsymbol{C}_t$ can be interpreted as a representation that encodes the functional connectivity among EEG channels.

\noindent\textbf{Problem (Manifold Representation Learning).}
Given a $\boldsymbol{X}_t$, we can compute its SPD representation via the sample covariance matrix:
$\boldsymbol{C}_t = \frac{1}{L} \sum_{\ell=1}^{L} (\boldsymbol{x}_{t,\ell} - \boldsymbol{\mu}_t)(\boldsymbol{x}_{t,\ell} - \boldsymbol{\mu}_t)^\top \in \mathbb{S}_{++}^N,$
where $\boldsymbol{x}_{t,\ell} \in \mathbb{R}^N$ is the EEG epoch at time $\ell$ in epoch $t$, and $\boldsymbol{\mu}_t$ is the channel-wise mean.
This symmetrically aggregates second-order statistics across all channel pairs, treating them equally without considering channel importance.
Meanwhile, the channel-wise mean assumes linear stationarity.
Since each covariance matrix is computed after mean-centering, the dynamics and temporal dependencies of neural signals might be flattened.
To tackle these challenges, we aim to integrate explicit graph connectivity
$\mathcal{A}_t \in \mathbb{R}^{N \times N}$ into the $t$-th covariance representation
$\mathcal{C}_t \in \mathbb{R}^{N \times N}$, thereby learning graph-aware and dynamic
SPD-manifold representations.
This integration raises the following key challenges.
\ding{182}\emph{Geometric incompatibility.}
Graph structures $\mathcal{A}_t$ are defined in Euclidean space,
whereas covariance matrices $\mathcal{C}_t$ lie on a curved Riemannian manifold.
How to design a principled integration mechanism that respects this geometric mismatch remains non-trivial.
\ding{183}\emph{Feature space alignment.}
How to align Euclidean graph features with manifold-valued covariance representations,
so as to ensure geometric consistency and construct a discriminative tangent space
suitable for downstream tasks.
We address each of these challenges through dedicated modules,
and validate their effectiveness via ablation studies.

\section{Methodology}
\label{sec:method}
Figure \@\ref{fig: proposal} illustrates the system overview of \method.
Specifically, Section~\ref{subsection:4.1} addresses
\ding{182}\emph{Geometric incompatibility}
by proposing a dynamic graph–guided modulation mechanism,
which leverages time-varying graph structures to compensate for the limited expressiveness
of static SPD representations.
Section~\ref{subsection:4.2} addresses
\ding{183}\emph{Feature space alignment}
via a geometry-aware feature space alignment module,
which ensures that the learned latent representations evolve consistently
within a Riemannian manifold while capturing dynamic structural information.

\subsection{Dynamic Graphs Guided SPD modulation}
\label{subsection:4.1}
%This subsection presents a dynamic graph guided integration strategy, where dynamic topological knowledge is injected into the statistical SPD representations. To do this, we first perform dimensional alignment between graph and SPD embeddings, followed by a manifold cross-attention designed to modulate SPD matrices with dynamic graph topology.

\paragraph{EEG SPD and Dynamic Graph Encoding.}
Given raw EEG signals and segmented epochs $\{\mathcal{X}_t\}_{t=1}^{T}$, we construct two complementary SPD sequences as inputs to cross-attention.
For the spatiotemporal view, we first extract features from the raw EEG signals and compute each EEG-view SPD matrix $\boldsymbol{S}_t\in\mathbb{S}_{++}^{d_c}$ following \cite{kobler2022spd}.
For the graph view, we propose a graph dynamic embedding using the time-then-graph framework to integrate a temporal graph structure \citep{kotoge2025evobrain}. This has been further enhanced by recent works that construct latent representations \cite{8237607,chen2023riemannian}.
Also we lift it to a spectral SPD matrix $\boldsymbol{C}_t \in\mathbb{S}_{++}^{d_c}$.
The resulting sequences $\{\boldsymbol{S}_t\}_{t=1}^{T}$ and $\{\boldsymbol{C}_t\}_{t=1}^{T}$ serve as the key/value and query inputs, respectively.
This forward process captures both the epoch variations between frequency bands and explicit channel correlations.

To enhance the local sensitivity of stochastic SPD representations to brain dynamics, we develop a reciprocal geometric-topology integration. This integration aims to leverage the structural priors encoded in the graph topology to enhance the geometric ability of SPD. Two key challenges arise:
First, graphs capture functional connectivity across spectral and EEG channels, providing explicit topological structure. However, given the geometric mismatch between the Euclidean nature of graph structures, due to their linear and vector operations, and the Riemannian nature of SPD manifolds. 
Moreover, aligning representations from dynamic graphs and SPD geometry requires a unified latent space where they can interact meaningfully. However, establishing such unified space is not sufficient. We further require dynamic geometric evaluation to adaptively calibrate the enhanced SPD, leading to more robust and structure-aware modeling of brain dynamics.

Motivated by \cite{gheini2021cross, pan2022matt}, we adopt the attention mechanism and define queries derived from dynamic graph embeddings as $\tilde{\mathcal{G}_{t}}$, with keys and values constructed from spatiotemporal information, enabling local topological and SPD modulation on the manifold. Unlike conventional dot-product attention in Euclidean space \cite{NIPS2017_3f5ee243}, it is invalid in a manifold. We measure the similarity between $\boldsymbol{K}$ and $\boldsymbol{Q}$ with Log-Euclidean geodesic distance on the SPD manifold, ensuring Riemannian consistency:
\[
        d_{\text{LE}}(\boldsymbol{K}_{i},\boldsymbol{Q}_{i}) =  \left\| \log(\boldsymbol{K}_{i}) - \log(\boldsymbol{Q}_{i}) \right\|_F.
\]

\paragraph{Manifold–Dynamics Graph Cross-Attention.}
To integrate the functional topology of dynamic graphs into stochastic SPD representations, we design a manifold-aware cross-attention mechanism.
Keys and values are derived from EEG-based latent SPD matrices, while queries are obtained from dynamic graph embeddings.
We obtain the key, value from SPD collection $\{\boldsymbol{S}_{t}\}_{t=1}^{T} \in \mathbb{S}^{d}_{++}$ and query from $\{\boldsymbol{C}\}_{t=1}^{T} \in \mathbb{S}^{d}_{++}$ through a bilinear mapping (BiMap) \cite{huang2017riemannian} to explore the nonlinear feature points in the manifold from each epoch: %Specifically, $\boldsymbol{K}_{i}$ and $\boldsymbol{V}_{i}$ are obtained through a bilinear mapping (BiMap) of $\boldsymbol{S}_i$ \cite{huang2017riemannian} to explore the nonlinear feature points in the manifold from each epoch. 
\begin{itemize}
    \item  $\boldsymbol{K}_{t} \in \mathbb{S}^{l}_{++}$ (\underline{Estimated SPD}) : \quad \quad \quad \quad $\boldsymbol{W}_{k}\boldsymbol{S}_{t}\boldsymbol{W}_{k}^{T}$,
    \item $\boldsymbol{V}_{t} \in \mathbb{S}^{l}_{++}$ (\underline{Estimated SPD}) : \quad \quad \quad \quad
$\boldsymbol{W}_{v}\boldsymbol{S}_{t}\boldsymbol{W}_{v}^{T}$, 
    \item  $\boldsymbol{Q}_{t} \in \mathbb{S}^{l}_{++}$ (\underline{Constructed graph SPD}): \quad $\boldsymbol{W}_{q}\boldsymbol{C}_{t} \boldsymbol{W}_{q}^{T}$ .
\end{itemize}
% We define the $\boldsymbol{S}_{i} \in \mathbb{S}^{d}_{++}$ as the $\boldsymbol{K}_{i}$, and $\boldsymbol{V}_{i}$ following the bilinear mapping (BiMap) \cite{huang2017riemannian} to explore the nonlinear feature points in the manifold from each epoch. 
%We convert the $\boldsymbol{C}_{i} \in \mathbb{S}^{d}_{++}$ to the $\boldsymbol{Q}_{i}$ also via BiMap operation. The mappings of $\boldsymbol{K}_{i}$, $\boldsymbol{Q}_{i}$, $\boldsymbol{V}_{i}$ from $\boldsymbol{C}_{i}$ and $\boldsymbol{S}_{i}$ are
%\begin{equation}
%\boldsymbol{K}_{i} = \boldsymbol{W}_{k}\boldsymbol{S}_{i}\boldsymbol{W}_{k}^{T}, \boldsymbol{Q}_{i} = \boldsymbol{W}_{q}\boldsymbol{C}_{i} \boldsymbol{W}_{q}^{T}, \boldsymbol{V}_{i} = \boldsymbol{W}_{v}\boldsymbol{S}_{i}\boldsymbol{W}_{v}^{T}, 
%\label{eq: bimap}
%\end{equation}
where $\boldsymbol{W}_{k}$, $\boldsymbol{W}_{q}$, and $\boldsymbol{W}_{v} \in \mathbb{R}^{d \times l}$ $(l < d )$ denote BiMap matrices. 
Note, the BiMap matrices are constrained as row-full rank, and the $\boldsymbol{K}_{t}, \boldsymbol{Q}_{t}, \boldsymbol{V}_{t} \in \mathbb{S}^{l}_{++}$ still live in the SPD manifold.
We construct keys and values from the spatiotemporal view SPD collection $\{\boldsymbol{S}_t\}_{t=1}^{T} \in \mathbb{S}^{d}_{++}$, as it inherently encodes the geometric configuration of brain states via second-order statistics and preserves spatial consistency under the Riemannian structure. 
Queries are constructed from a dynamic graph SPD collection $\{\boldsymbol{C}_{t}\}_{t=1}^{T} \in \mathbb{S}^{d}_{++}$, as it explicitly encodes functional connectivity and provides structures to selectively guide the integration of brain representations on the manifold. 

%Once we obtained the $\boldsymbol{K}_{i}$,$\boldsymbol{Q}_{i}$, and $\boldsymbol{V}_{i}$, we enhance the local topological awareness ability of the SPD manifold by measuring the domain similarity between $\boldsymbol{K}_{i}$ of the spatiotemporal and $\boldsymbol{Q}_{i}$ of the dynamic graph. This cross-domain integration of $\boldsymbol{K}_{i}$ ,$\boldsymbol{Q}_{i}$ typically follows the procedure of cross attention with the dot product \cite{NIPS2017_3f5ee243} in Euclidean space. 

We establish a unified space to integrate the topology into SPD representation and calibrate the enhanced SPD based on the manifold cross-attention, leading to more robust and structure-aware modeling of brain dynamics.
The attention operation is transformed via Log-Euclidean geodesic distance computation in the case of the SPD manifold. 
The manifold cross-attention score measures the geometric affinity between the graph-induced query $\boldsymbol{Q}_j$ and the SPD key $\boldsymbol{K}_t$ via the Log-Euclidean distance:
\begin{equation}
    \mathcal{A}_{t,j}= \frac{1}{1+\log(\mathbf{1}+d_{\text{LE}}(\boldsymbol{K}_t,\boldsymbol{Q}_j))}  .
    \label{eq: corss-atten}
\end{equation}

Unlike standard Transformer attention, we normalize the cross-attention scores with a row-wise softmax function.
Therefore, each graph-guided query selectively attends to relevant SPD representations across time:

\begin{equation}
    a_{t,j} = \frac{\exp(\mathcal{A}_{t,j}/\tau)}{\sum_{k=1}^{T}\exp{(\mathcal{A}_{t,k}/\tau})} .
    \label{eq: prob ca}
\end{equation}

Finally, the attention weights in Eq.~\ref{eq: prob ca} are used to aggregate $\boldsymbol{V}_{t}$ SPD matrices, allowing graph-guided selection of meaningful geometric features via a Log-Euclidean weighted mean:
The enhanced SPD representation is obtained by mapping the weighted combination back to the manifold via the exponential map:
\begin{equation}
    \tilde{\boldsymbol{S}}_t = \exp{\sum_{j=1}^{T}(a_{t,j}\log(\boldsymbol{V}_{j}))}  .
\end{equation}

This setup naturally enables topological guidance over SPD geometry across time, as all representations are segment-specific and temporally indexed by epoch $t$.
Hence, dynamic graph-induced spectral synchrony and channel-wise connectivity can be integrated SPD representations.

\subsection{Geometry-Aware Feature Space Alignment}
\label{subsection:4.2}
After the dynamic graphs guided SPD modulation, we extract the geometric features of the modulated SPD $\tilde{\boldsymbol{S}}_{t} \in \mathbb{S}^{l}_{++}$ with ReEig (filter the eigenvalues of $\tilde{\boldsymbol{S}}_{t}$) and obtain the tangent embedding  ${\boldsymbol{M}}_{t}$ with $\log$ operation $\mathcal{M} \mapsto\mathcal{T}_{p}\mathcal{M} $ for the downstream task. The $\log$ operation is a common technique in SPD learning \cite{huang2017riemannian} to project SPD data to the Euclidean space, allowing the classifier to work in a flat space. We define the $\log$ operation as $\psi: \mathbb{S}^{l}_{++} \mapsto \mathbb{R}^{(l \times l+1)/2}$:
\[
   {\boldsymbol{M}}_{t} = \psi(\tilde{\boldsymbol{S}}_{t})= \boldsymbol{U}(\log(\sigma_{1}), \dots,\log(\sigma_{l}))\boldsymbol{U}^{T} ,
\]
where $\boldsymbol{U}$ is the eigenvector matrix of $\tilde{\boldsymbol{S}}_{t}$, and $\sigma$ are the eigenvalues. We then apply a fully connected layer to  the flattened ${\{\boldsymbol{M}}\}_{t=1}^{T}$ followed by a  regular \texttt{softmax} in the tangent space. We obtain the final output $\hat{\boldsymbol{y}}$, and measure with the ground truth $\boldsymbol{y}$ using the cross-entropy loss $\mathcal{L}_\text{CE}$.

To suppress graph drift and stabilize the discriminative structure in tangent space, we propose a geometry-aware loss $\mathcal{L}_\text{GeoTop}$ to avoid a dimensional deformation. 
% 对于不同视角观测下的几何交互 - 共享什么好处？
% 对于切空间上的决策boundary 有什么好处？ 
% 我们关注的不再是分的对，而是怎么能让他分的对。
%\begin{itemize}
%    \item \textbf{Representation-level objective}. The SPD geometric and topological consistency supervisory signals impose alignment constraints on both graph and SPD branches at the loss level.
%    This is to further enhance the information complementarity and facilitate fusion of graph and SPD in modulation.    
%    \item \textbf{Decision-level objective}. 
%    Structural consistency constraints enable SPD representations to be more discriminative in the tangent space, reshaping the discriminative boundaries by pulling the same-attribute geometric information closer and pushing the irrelevant away.
%\end{itemize}
%we are more concerned about know-how than numerical results. 
Specifically, our framework regards the graph topology and the SPD geometric embedding in tangent space (via $\log$ projection) as two complementary views of the raw EEG signals: the former captures the explicit functional structure among brain regions, while the statistical geometry. Aligning these two representations parallels the objective of representation learning, i.e. enforcing semantic consistency across heterogeneous views, which allows the model to extract more robust and informative representations for downstream tasks \cite{hassani2020contrastive, shen2022contrastive}. The proposed $\mathcal{L}_\text{GeoTop}$ is formulated as
\begin{equation}
    \mathcal{L}_\text{GeoTop} = -\frac{1}{N}\sum_{i=1}^{N}\log
\frac{\exp(\mathrm{sim}(h(\boldsymbol{g}_i),h(\boldsymbol{u}_i))/\kappa)}
{\sum_{j=1}^{N}\exp(\mathrm{sim}(h(\boldsymbol{g}_i),h(\boldsymbol{u}_j))/\kappa)},
\end{equation}
where $\kappa$ is a temperature parameter, $\mathrm{sim}(\cdot)$ is a cosine correlation to align the pooled tangent embedding $\boldsymbol{u}_i$ with the global graph feature $\boldsymbol{g}_i$ by feedforward projection $h$. %in Eq. \ref{eq: global_g}.
The $\mathcal{L}_\text{GeoTop}$ encourages the SPD embeddings to be closer to the corresponding graph structures while repelling irrelevant ones.
Finally, the total loss $\mathcal{L}$ is
\begin{equation}
    \mathcal{L}=\mathcal{L}_\text{CE}+\beta \mathcal{L}_\text{GeoTop} ,
\end{equation}
where $0<\beta<1$ is a hyperparameter to control the contributions of the two space alignment.
In summary, $\mathcal{L}_\text{GeoTop}$ offers a geometric-aware regularization that enhances the SPD modulation pathway, supporting a grounded learning process through the manifold cross-attention mechanism.

% 用实验如何证明/分析 论证spd的统计特性 不能很好的在流形上不能很好的表达dynamic
% BCI的实验结果，subject和session 的结果区分好
% 复杂度分析
% attention机制的特征融合的有效性，
% 也可以从反推的角度 分析data-driven的SPD方法，ie，matt虽然引入了attention的计算机制，一定程度上达到了xxx样的效果，但是xxxxx。
% matt 理论上是不是堆叠attention的layers，即越多应该表现会好呢？我们发现增加了attention 层数不能更好的增加spd attention的表达能力 （这个结论要是有实验论证是最好的，比如可视化 初始SPD 和经过6层之后的SPD matrix。我们发现会削弱spd的统计特性 -> 趋同）
\section{Experiments}

In this section, we organize experiments to address the following research questions
derived from the challenges outlined in Section~\pageref{sec:problem}:\\
\textbf{RQ1.} Does \method enhance the representational capacity of SPD matrices
by introducing geometry-aware cross-attention on EEG datasets?
\\
\textbf{RQ2.} How does the proposed geometric formulation influence the construction
of a temporally and spatially consistent manifold cross-attention metric?
\\
\textbf{RQ3.} Does the geometric coupling objective facilitate stable and effective optimization
on the SPD manifold $\mathbb{S}_{++}^{n}$?

% In this section, we organize the experiments to answer the following research challenges in section~\pageref{sec: problem}:  
% \begin{itemize}[left =0pt]
% \item \textbf{Accuracy of forming the manifold cross-attention.} \quad
% Does \method enhance the capacity of SPD by introducing the geometric function on EEG datasets?

% \item \textbf{Addressing temporal-spatial consistent graph space.} \quad 
% How does the property of geometric function affect the establishment of manifold cross-attention metric? 
% \item \textbf{Meaningful objective Learning.} \quad
% Does the geometric coupling objective facilitate  optimization on $\mathbb{S}_{++}^{n}$?
% \end{itemize}

We first introduce the task settings and the experimental dataset applied for evaluation, followed by detailed results and analysis to answer each of the above questions.
We also conduct efficiency analysis and ablation studies.

\subsection{Experimental Setup}
\paragraph{Tasks and Datasets.} In this study, we evaluated \method, enhancing the SPD manifold via dynamic graphs, with the seizure detection and motor imagery (MI) classification. 
Seizure detection is a binary classification task that distinguishes between seizure and non-seizure EEG epochs, serving as a core component in automated seizure monitoring systems.
MI full classification task in BCI applications, aiming to decode user intent based on imagined movements.

For seizure detection, we used the Temple University Hospital EEG Seizure dataset v1.5.2 (TUSZ) \cite{shah2018temple}, the largest publicly EEG seizure database, to evaluate our \method. TUSZ contains 5,612 EEG recordings with 3,050 annotated seizures. Each recording consists of 19 EEG channels following the 10-20 system, ensuring clinical relevance. A key strength of TUSZ lies in its diversity, as the dataset includes data collected over different time periods, using various equipment.

For MI, we utilized the BCIC-\uppercase\expandafter{\romannumeral4}-2a as a representative benchmark featuring time-asynchronous EEG data. The BCIC-\uppercase\expandafter{\romannumeral4}-2a dataset is one of the most widely used public EEG dataset released for the BCI Competition IV in 2008 \cite{brunner2008bci}.
It contains EEG recordings in a four-class MI task from nine subjects, each participating in two sessions conducted on different days. In each trial, subjects were cued to imagine one of four movements—right hand, left hand, feet, and tongue for four seconds after an instructional cue. Each session includes 288 trials (i.e., 72 trials per class). EEG signals were recorded using 22 Ag/AgCl electrodes over the central and surrounding regions at a sampling rate of 250 Hz.
We conduct experiments on the publicly available BNCI 2014-001 dataset from the MOABB benchmark suite \cite{Aristimunha_Mother_of_all}. 
We apply standard preprocessing to the 22-channel EEG signals, including: (1) down-sampling (256 Hz to 200 Hz), (2) band-pass filtering between 4–38 Hz, and (3) trial segmentation by extracting the 0.5–4.0s interval after cue onset for each trial.

\paragraph{Baseline methods.}
For seizure detection task, we selected the previous and current SOTA models for comparison with \method, including: DCRNN \cite{li2017diffusion}, which performs dynamic graph modeling with the general Euclidean metric. We compare with a Transformer baseline, BIOT \cite{yang2023biot}, which captures temporal-spatial information for various EEG tasks. We also evaluate temporal-spatial CNN-LSTM-based model \cite{9175641}. For DL-based SPD models, we compare with the SPD learning-based SPDNet \cite{huang2017riemannian}, and self-attention-based MAtt \cite{pan2022matt}. \\
For MI classification, we selected previous and current SOTA models for comparison with \method, including: parallel CNN blocks-based FBCNet \cite{mane2021fbcnet}, and SPD matrix-learning-based SPDNet \cite{huang2017riemannian}. Tensor-CSPNet \cite{ju2022tensor} performs spectral SPDNet from multi-frequency banks. Graph-CSPNet \cite{ju2022graph} offers a graph structure-based SPD learning view to improve efficiency. 
MAtt \cite{pan2022matt} performs a self-attention in SPD manifold with the local EEG segments. 
CorAtt \cite{hu2025ijcai} offers a robust SPD attention with the full-rank correlation. 
%ShallowConvNet\cite{kollHod2023deep} applies a CNN as feature extraction and combines a Riemannian-based discriminator. DeepConvNet \cite{schirrmeister2017deep} is a enhanced version of ShallowConvNet with deeper CNN encoder. MDM \cite{zhang2020manifold} performs common spatial pattern (CSP) and Riemannian-based discriminator. FgMDM\cite{barachant2010riemannian} is a SPD covariance-based model.

\paragraph{Metrics and Training Settings.}
We evaluated the model using the Area Under the Receiver Operating Characteristic curve (AUROC) and the F1 score. AUROC measures the ability of models across varying thresholds, while the F1 score highlights the balance between precision and recall at its optimal threshold for the classification task.

Training for all models was optimized using the Adam optimizer \cite{kingma2014adam} with an initial learning rate of $1 \times 10^{-3}$ in the PyTorch and PyTorch Geometric libraries on NVIDIA A6000 GPU and AMD EPYC 7302 CPU. During manifold training, we applied the meta-optimizer to ensure the mathematical consistency between the Riemanmian geometry and the general encoder model.

\begin{figure}[t]
\centering
\includegraphics[width = 0.99\linewidth]{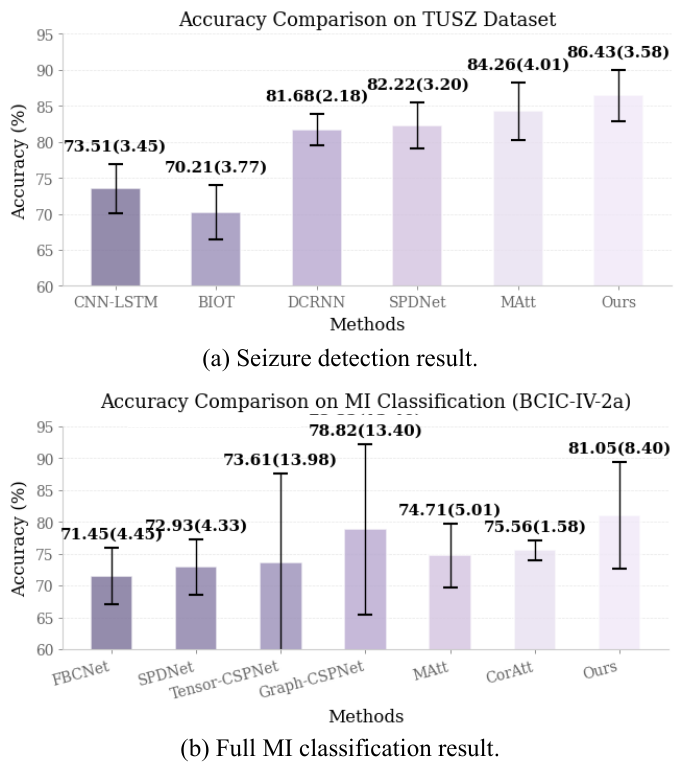}
\caption{Accuracy comparison on seizure detection and MI full classification. Bars show the mean accuracy and error bars indicate the standard deviation across 5-fold runs.
}
\label{res:main_acc}
\end{figure}

\subsection{Experimental Results}

\paragraph{Effectiveness of manifold cross-attention.}
To answer \textbf{RQ1}, we first present the seizure detection results in Figure~\ref{res:main_acc}(a) within 12s EEG signals. Traditional models like CNN-LSTM and BIOT exhibit limited generalization capability, with lower accuracy scores. The graph-based method DCRNN demonstrates a significant performance boost by leveraging dynamic spatial dependencies, achieving $81.68\%$. Manifold-based models (SPDNet and MAtt) further explore the intrinsic Riemannian structure of EEG representations, where MAtt achieves a higher score than SPDNet ($82.22\%$ to $86.43\%$).
Our proposed \method model outperforms all baselines, achieving the best score (86.43\%). These results validate the benefit of jointly integrating functional connectivity and SPD geometric features through manifold cross-attention modulation and structure-aware alignment, leading to improved seizure detection and generalization.

%\begin{table}[t] 
%\centering
%\caption{Performance comparison on BNCI2014001 dataset for all classification. \textbf{Bold} and \underline{underline} indicate best and second-best results.}
%\begin{NiceTabular}{ll} 
%    \toprule[1.2pt]
%       \rowcolor{gray!20}
%Method & ACC \\
%       \midrule
%FBCNet & 71.45 \\
%SPDNet & 72.93 \\
%Tensor-CSPNet &  73.61  \\
%Graph-CSPNet &  \underline{78.82} \\
%MAtt   &  74.71 \\
%CorAtt & 75.56 \\
%\midrule
%\method &  \textbf{81.05}  ($\uparrow \textbf{3.76}$) \\
%\bottomrule
%\end{NiceTabular}
%\label{tab:MI}
%\end{table}

Figure~\ref{res:main_acc}(b) represents the performance of different models on the BNCI2014001 dataset for motor imagery classification. On BCIC-IV-2a MI classification, our method achieves the highest accuracy of 81.05\% . It outperforms the strongest baseline Graph-CSPNet (78.82\%) in absolute accuracy. Compared with other attention/manifold baselines, our method yields clear gains over CorAtt (75.56\%) and MAtt (74.71\%). Traditional baselines such as Tensor-CSPNet, SPDNet , and FBCNet are further behind. Overall, these results suggest that incorporating dynamic structural priors with geometry-consistent SPD fusion improves discriminability for MI classification. Notably, \method shows much lower performance variance, suggesting greater robustness and stability. This result demonstrates that our SPD-based geometric modulation captures both global and local connectivity patterns more effectively, leading to superior discriminative representation under low-data regimes.

\begin{figure}[t]
\centering
\includegraphics[width = 0.99\linewidth]{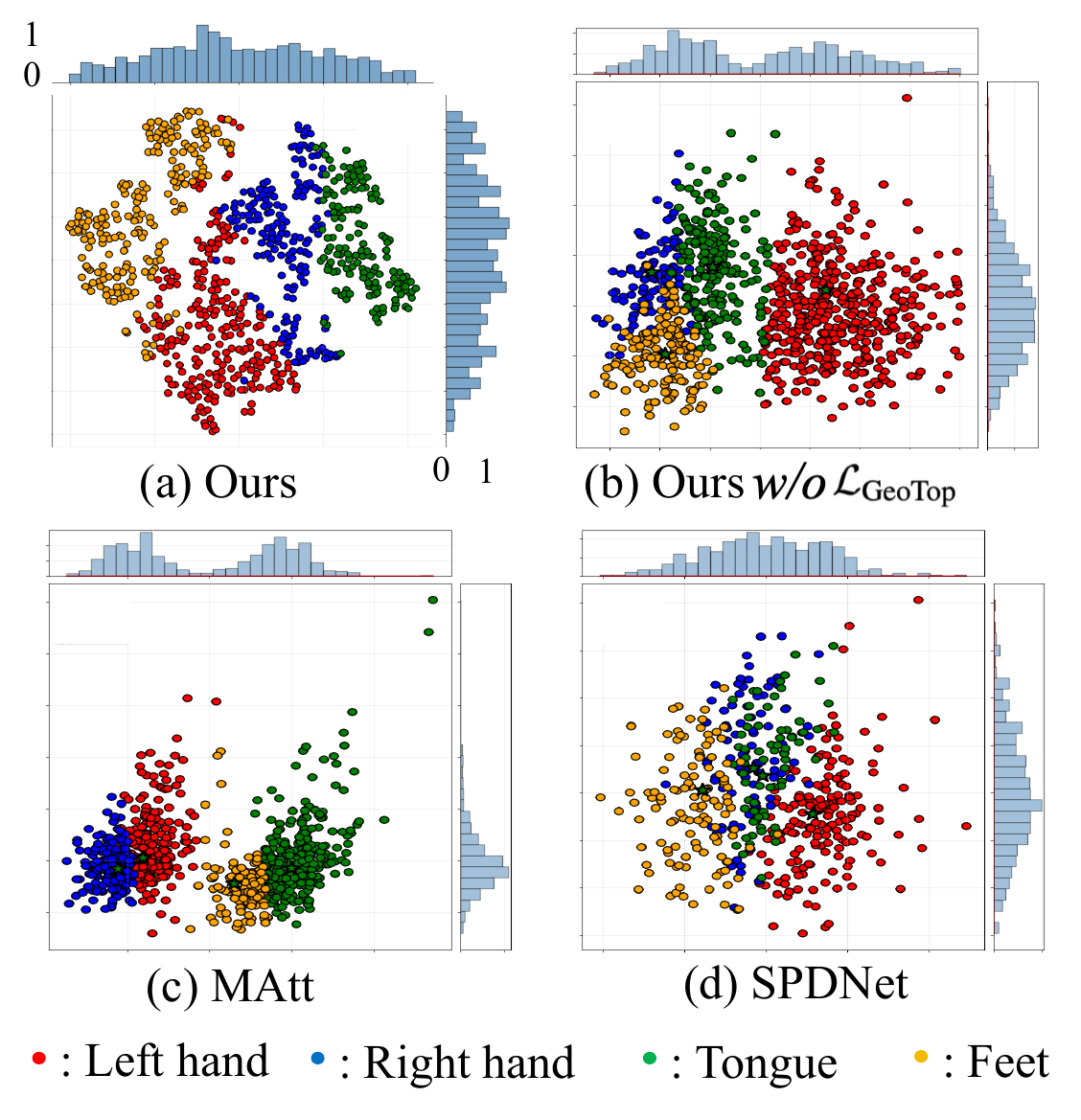}
\caption{Visualization of tangent-space with 2D PCA. (a) \method produces well-structured clusters. (b) w/o $\mathcal{L}_{GeoTop}$ leads to increased overlap. (c) MAtt captures spatiotemporal patterns via manifold attention but yields looser clusters. (d) SPDNet exhibits the mixing clusters by the second-order statistics.}
\label{res:PCA}
\end{figure}

\begin{figure}[t]
\centering
\includegraphics[width = 0.95\linewidth]{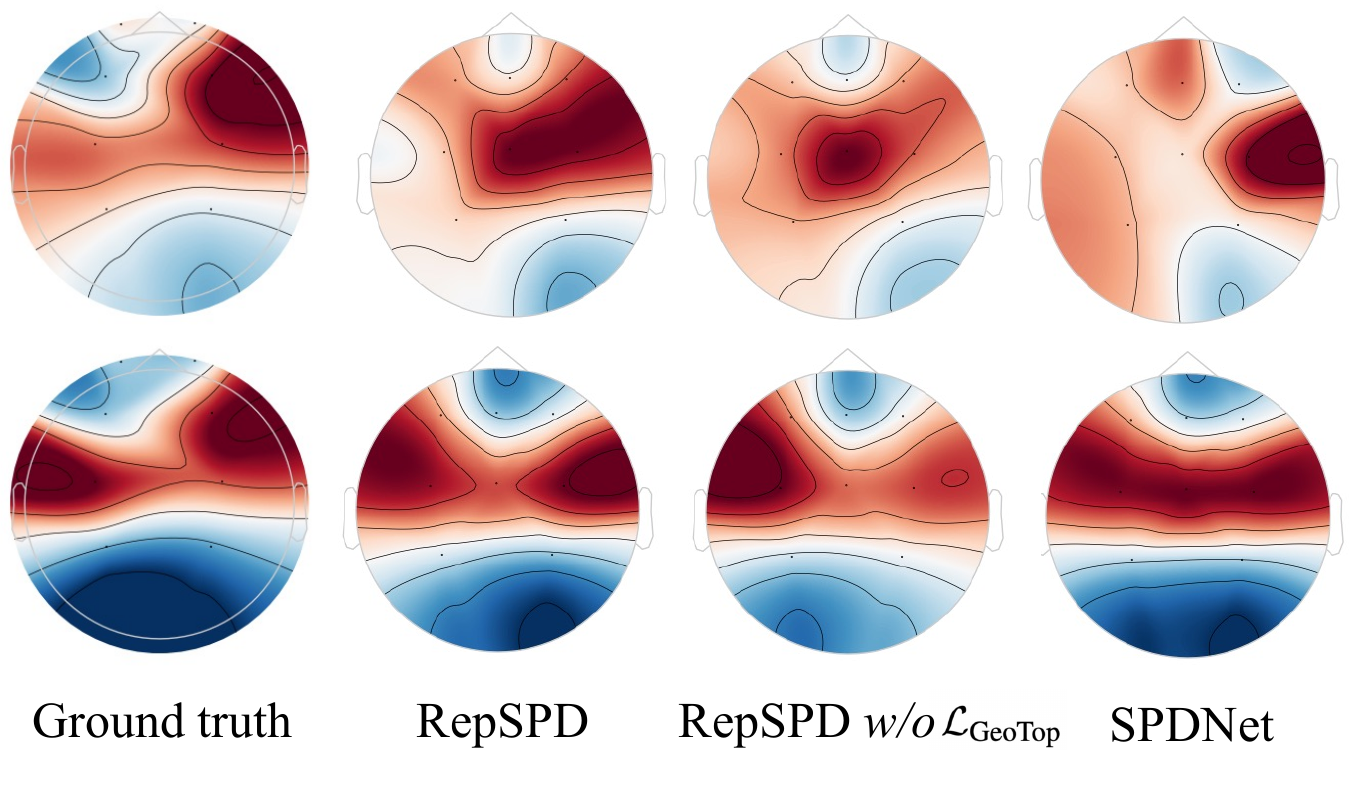}
\caption{Visualization of topological map of brain across different methods. Compared to SPDNet and \method w/o proposed $\mathcal{L}_{GeoTop}$, our \method more accurately restores the functional spatial pattern observed in the ground truth.
%In the first row, \method captures a broader high-activation area that shifts toward the upper-right region, closely matching the true distribution. In the second row, \method clearly reconstructs the bilateral contrast between left and right hemispheres, revealing sharper topological boundaries and more faithful spatial gradients.
}
\label{res: spatial}
\end{figure}

\begin{figure}[t]
\centering
\includegraphics[width = 0.99\linewidth]{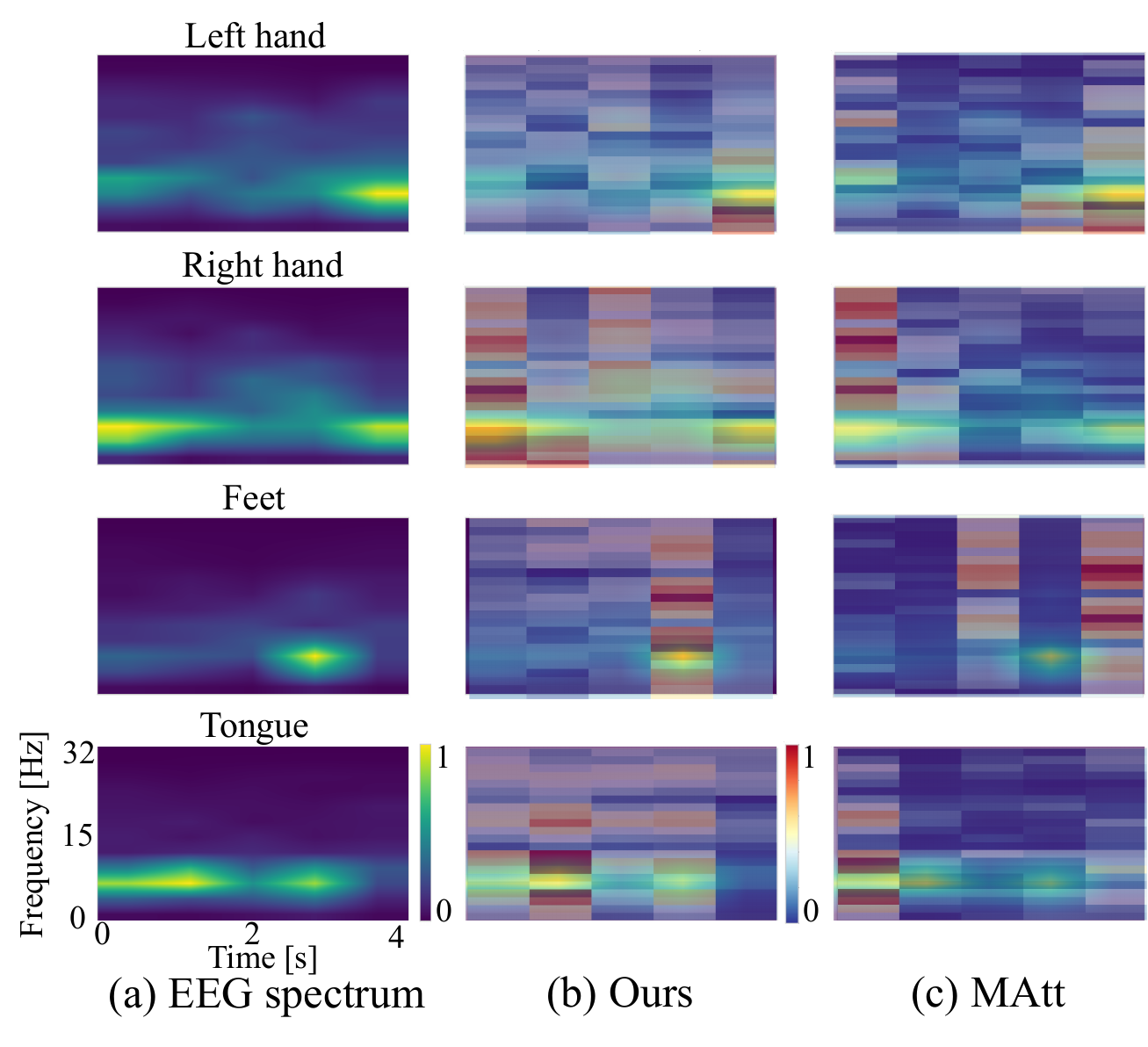}
\caption{Gradient-level visualization of attention maps. (a) Spectrogram of input EEGs for each class. (b) \method produces more structured and dynamic attention patterns that emphasize time–frequency regions. (c) MAtt yields less class-specific attention. 
}
\label{res:grad}
\end{figure}

\paragraph{Temporal--spatial consistency in geometry.} To answer the \textbf{RQ2}, Figure~\ref{res:PCA} shows the 2D projection of the learned tangent-space to compare the effects of the graph-structured prior and $\mathcal{L}_{GeoTop}$ for stabilizing the discriminative geometry across the existing methods.
(d) SPDNet relies on second-order SPD statistics, which tend to smooth out time-varying connectivity, leading to the most class overlap.
(c) MAtt introduces local SPD representations and manifold attention to capture spatiotemporal patterns, yet the absence of an explicit structural prior results in relatively loose clusters.
(b) The dynamic graph prior helps encode cross-channel dependencies and time-varying connectivity by our SPD cross-attention,  improving inter-class separation. However, without cross-view calibration, it still causes noticeable overlap.
(a) \method yields more compact and well-separated clusters, indicating that geometry-consistent fusion together with cross-view alignment improves the discriminability of tangent-space embeddings.

% Figure~\ref{res: spatial} illustrates the spatial pattern comparison across different methods. The proposed \method with contrastive loss shows a distribution that best approximates the ground truth, while “w/o loss” and SPDNet exhibit spatial drift and local ambiguity. The result highlights the effectiveness of our proposed loss, which—beyond task-driven optimization—encourages geometric reciprocity between structural and discriminative learning objectives. By introducing structure-aware supervision at the geometric level, our model enhances the modulation of SPD representations, leading to more interpretable and discriminative projections in the tangent space.

Figure~\ref{res: spatial} visualizes the recovered spatial brain patterns across different methods.
Compared with the ground truth, the proposed \method with the contrastive geometric loss produces a spatial distribution that most closely preserves the global functional topology as well as local activation contrasts.
In contrast, \method without the proposed loss exhibits noticeable spatial drift, where functionally related regions become blurred or misaligned, while SPDNet shows fragmented patterns with increased local ambiguity.
These differences highlight the role of the proposed contrastive loss in regularizing representation learning beyond task-driven supervision.
Rather than optimizing only for discriminative performance, the loss explicitly enforces geometric consistency between learned SPD representations and the underlying spatial topology of brain activity.
By introducing structure-aware supervision at the geometric level, our model better modulates the embedding of SPD matrices.

\paragraph{Objective Learning effectiveness.} To answer \textbf{RQ3}, Figure~\ref{res:grad} illustrates the gradient-level interpretability of attention maps over time and frequency-band tokens. The reference time–frequency patterns are computed from the input EEGs using log-spectral magnitudes. Compared with MAtt, \method yields more structured and class-dependent attention distributions, placing higher weights on discriminative regions while suppressing irrelevant effects. The behavior of attention maps are consistent with the improved clustering in tangent space (Figure~\ref{res:PCA}), suggesting that injecting dynamic structural priors and performing geometry-consistent alignment stabilizes representations from token selection to downstream decision.

\paragraph{Ablation Study.}
We consider two ablation studies to evaluate the dynamic graph-based SPD modulation and geometric consistency loss design in our \method.
Table~\ref{tab: query abl} describes the effect of different query designs in the manifold cross-attention mechanism. Using either edge-based or node-based SPD queries yields strong performance, with node SPD achieving the best results (ACC: 86.43, AUROC: 81.61, F1: 42.92). The graph priors provide more direct and discriminative functional connectivity for guiding topology-aware modulation. In contrast, mixed queries lead to slight performance degradation, suggesting that overly mixing geometric cues may introduce redundancy or interfere with geometry alignment. This validates that domain-consistent SPD query design is critical for capturing interpretable and discriminative tangent space.

\begin{table}[t] 
\centering
\caption{Ablation comparison with different query options, including edge-based, node-based, node and edge mixed-based, and node-EEG SPD mixed queries.
}
\begin{NiceTabular}
{l c c c} 
    \toprule[1.2pt]
       \rowcolor{gray!20}
Query input & ACC & AUROC & F1 \\
       \midrule
edge SPD & \underline{86.13} & \textbf{81.66} & \underline{41.88} \\%
node SPD & \textbf{86.43} & \textbf{81.61} & \textbf{42.92} \\
edge and node mixed SPD & 85.17 & 80.83 & 41.44\\
node and $S$ mixed SPD & 85.22& 80.47& 40.42\\
\bottomrule
\end{NiceTabular}
\label{tab: query abl}
\end{table}

\begin{table}[htbp] 
\centering
\caption{Ablation comparison with geometry-aware feature space alignment loss $\mathcal{L}_{\text{GeoTop}}$.
}
\begin{NiceTabular}
    {l c c c} 
    \toprule[1.2pt]
    \rowcolor{gray!20}
$\lambda$ & ACC & AUROC & F1 \\
       \midrule
0 (w/o $\mathcal{L}_{\text{GeoTop}}$ ) & 77.81 & 79.57 & 37.80 \\
0.1  & 80.52 & 76.61 & 36.90 \\
0.3   & \textbf{86.43} & \textbf{81.61} & \textbf{42.92} \\
0.5   & \underline{82.31} & \underline{81.47} & \underline{39.12} \\
0.7   & 76.87 & 79.13 & 38.92 \\
\bottomrule
\end{NiceTabular}
\label{tab: nce abl}
\end{table}

%对于cross attention调制的引导作用，以及对决策边界的作用
Table~\ref{tab: nce abl} evaluates the contribution of our proposed geometry-aware contrastive loss. Removing this loss significantly reduces all metrics (e.g., F1 from 42.92 → 37.80), highlighting the effectiveness of our cross-view SPD-tangent alignment strategy. By encouraging geometric consistency between SPD-based and graph-based embeddings, the model learns a more robust and distinguishable representation.
These gains indicate that cross-view geometric-alignment helps stabilize the discriminative structure in tangent space.
On the other hand, we select a small $\lambda$, the alignment is insufficient, leading to limited improvements and a slight drop in F1. Conversely, large $\lambda$ degrades performance, suggesting that excessive alignment can dominate the task objective and lead to over-regularize the representation.

\section{Conclusion}
We proposed \method, a novel geometric deep learning framework that integrated graph-based functional connectivity into the SPD manifold for EEG modeling. Unlike conventional approaches that treat SPD matrices as isolated geometric entities derived from statistical aggregation, our method introduced a cross-attention mechanism to modulate SPD features with graph-induced local topological priors. 
We further proposed a geometry-topology consistency loss to align SPD representations with global graph embeddings, enhancing their discriminability in the tangent space.\\
We acknowledge several directions for future work.
In the graph learning phase, \method relies on initially estimated inter-channel similarities to construct dynamic graphs; however, these graph structures remain fixed during training.
Future work will explore joint structure optimization to enable adaptive graph learning.

\bibliographystyle{icml2026}
\bibliography{example_paper}

%%%%%%%%%%%%%%%%%%%%%%%%%%%%%%%%%%%%%%%%%%%%%%%%%%%%%%%%%%%%%%%%%%%%%%%%%%%%%%%
%%%%%%%%%%%%%%%%%%%%%%%%%%%%%%%%%%%%%%%%%%%%%%%%%%%%%%%%%%%%%%%%%%%%%%%%%%%%%%%
% APPENDIX
%%%%%%%%%%%%%%%%%%%%%%%%%%%%%%%%%%%%%%%%%%%%%%%%%%%%%%%%%%%%%%%%%%%%%%%%%%%%%%%
%%%%%%%%%%%%%%%%%%%%%%%%%%%%%%%%%%%%%%%%%%%%%%%%%%%%%%%%%%%%%%%%%%%%%%%%%%%%%%%
\newpage

\appendix

\onecolumn
\section*{Appendix}
This appendix provides additional mathematical details of the manifold cross-attention mechanism, Riemannian backpropagation, implementation details, dataset descriptions, and supplementary experimental results. These contents aim to enhance the transparency and reproducibility of \method.
\section{Dynamic Spectral Graph Structure}\label{sec:graph}

% EEG signals are high-dimensional, multi-channel and have complicated space-time-frequency characteristics, making it difficult to capture real neural dynamics. 
Raw EEG signals consist of complicated neural activities overlapping in multiple frequency bands, each potentially encoding different functional neural dynamics. 
Directly analyzing EEG signals in the time domain often misses subtle state transitions occurring uniquely within specific frequency bands \citep{yang2022unsupervised}. 
Hence, it is beneficial to represent the intensity variations of frequency bands and waveforms by decomposing raw EEG signals into frequency components.
To effectively provide detailed insights for subtle state transitions, we perform the short-time Fourier transform (STFT) to each EEG epoch, preserving their non-negative log-spectral. Consequently, the multi-channel EEG recordings are processed as:
\begin{equation}
     \mathbf{X}_{t} = \sum_{t= \infty}^{-\infty}x[t] \,\omega[t-m]e^{-jwt},
\end{equation}
and a sequence of EEG epochs with their spectral representation is formulated as $\mathbf{X} \in \mathbb{R}^{N \times d \times T}$.

We then apply a graph representation by measuring the similarity among spectral representation $\mathbf{X}$ across EEG channels. Specifically, we define an adjacency matrix $\mathcal{A}_{t}(i,j)$ at each epoch $t$ as follows:
$\mathcal{A}_{t}(i,j) = \text{sim}(\mathbf{X}_{i,t},\mathbf{X}_{j,t})$
 and compute the normalized correlation between nodes $v_{i}$ and $v_{j}$, where the graph structure and its associated edge weight matrix $A_{i,j}$ are inferred from $X_{t}$ on for each $t$-th epoch. 
 We only preserve the top-$\tau$ highest correlations to construct the evident graphs without redundancy. To avoid redundant connections and clearly represent dominant spatial structures, we retain only the top-$\tau$ strongest connections at each epoch for sparse and meaningful graph representations.
Thus, we obtain a temporal sequence of EEG spectral graphs $\{{G}_t=(\mathcal{V}_t,\mathcal{A}_t)\}_{t=0}^{T}$.

\textbf{Temporal Graph Representation.}
Consider an EEG $\mathbf{X}$ consisting of $N$ channels and $T$ time points, we represent $\mathbf{X}$ as a graph, denoted as $\mathcal{G} = \{ \mathcal{V}, \mathcal{A}, \mathbf{X} \}$, where $\mathcal{V} = \{ v_1, \dots, v_N \}$ represents the set of nodes. 
Each node corresponds to an EEG channel. 
The adjacency matrix $\mathcal{A} \in \mathbb{R}^{N \times N \times T}$ encodes the connectivity between these nodes over time, with each element $a_{i,j,t}$ indicating the strength of connectivity between nodes $v_i$ and $v_j$ at the time point $t$.
Here, we redefine $T$ as a sequence of EEG segments, termed ``epochs'', obtained using a moving window approach. 
% If nodes $v_i$ and $v_j$ are connected by an edge $e_{i,j}$, the corresponding entry $a_{i,j,t}$ in the adjacency matrix is non-zero, reflecting their connectivity at $t$-th epoch.
The embedding of node $v_i$ at the $t$-th epoch is represented as $h_{i,t} \in \mathbb{R}^m$.
Specifically, we perform the short-time Fourier transform (STFT) to each EEG epoch, referring to \citep{GNN_ICLR22}.
Then we measure the similarity among the spectral representation of the EEG channels to initial the $\mathcal{A}_{t}(i,j)$ for each epoch $t$.

\subsection{Dynamic Graph Embedding Pipelines}
We perform the graph dynamic embedding to integrate a temporal graph structure to $\boldsymbol{S}$ \citep{gao2021equivalence}.
The graphic perspective provides an explicit method to formulate brain functional connectivity \cite{cui2022braingb}. 
To effectively capture such graph dynamics, we employ a graph dynamic embedding, following \textit{time-then-graph} framework to conduct the temporal graph structure, referring to \cite{gao2021equivalence}.

Practically, given a sequence of functional connectivity graphs $\mathcal{G}$, we first independently represent the nodes and edges, followed by embedding.
Specifically, given the $\mathcal{G}$ sequence as input, we first perform the sequence representation of the node and edge attributes independently.
Given the spectral attribute sequences $\{\mathcal{V}_{i,t} \}_{t =1 }^{T}$ of node/edge $i$ with intensity $ \{\mathcal{X}_{i,t}\}_{t=1}^{T}$, the node/edge evolution $\boldsymbol{h}_{i}^{n}$ with GRU is:
For node-level embedding, given the spectral attribute sequences $\{\mathcal{V}_{i,t} \}_{t =1 }^{T}$ of node $i$ with corresponding spectral intensity $ \{\mathcal{X}_{i,t}\}_{t=1}^{T}$, the node evolution $\boldsymbol{h}_{i}^{n}$ with GRU described as:
\[
\boldsymbol{h}_{i,t}^{n} = \text{GRU}^\text{node}(\mathcal{X}_{i,t}), \quad \boldsymbol{h}_{ij,t}^{e}=\text{GRU}^\text{edge}(\mathcal{A}_{ij,t}).
\]
%% check the math 
where the GRU explicitly learns the epoch-dependent evolution for each node. Similarly, we apply another GRU to model edge-level attribute sequences defined from adjacency matrices:
Finally, the resulting node and edge embeddings are  integrated to output dynamic graph structures $\tilde{\mathcal{G}_{t}} = (\boldsymbol{h}^{n}_{i,t},\boldsymbol{h}_{ij,t}^{e})$ and the graph neural network %(GNN) captures the spatial dependencies across the epochs $0:t$:
\begin{equation}
\boldsymbol{u}_{i}=\text{GNN}\left(\boldsymbol{h}_{i}^{n},\boldsymbol{h}_{ij}^{e}\right) .
\label{eq: global_g}
\end{equation}
This forward process captures both the epoch variations between frequency bands and explicit channel correlations.

\section{Manifold Cross-attention Pipelines}
\begin{figure}[t]
\centering
\includegraphics[width = 0.59\linewidth]{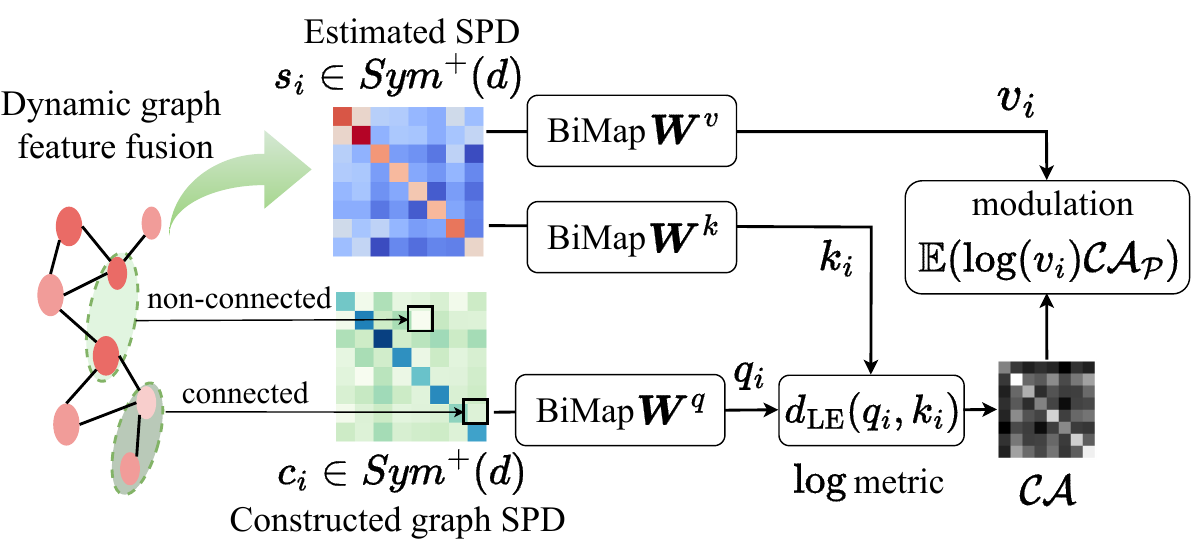}
\caption{Dynamic graph embedding-based SPD modulation via the proposed manifold cross-attention.
Given the estimated SPD matrix $s_i$ and graph-constructed SPD matrix $c_i$, BiMap projections yield query $(q_i)$, key $(k_i)$, and value $(v_i)$ representations. 
A Log-Euclidean metric computes attention scores $d_{\mathrm{LE}}(q_i, k_i)$, which modulate the value embedding in SPD space through $\mathbb{E}[\log(v_i)\mathcal{C}_\mathcal{A}]$, enabling geometry-aware feature integration.}
\label{fig:graph_spd}
\end{figure}

There exists a geometric mismatch when transitioning from Euclidean graph representations to SPD matrices defined on a Riemannian manifold. To address this discrepancy, we consider the Log-Euclidean distance as an alternative to the traditional Euclidean dot product. Figure~\ref{fig:graph_spd} shows the pipeline of proposed SPD corss-attention. Due to the abilities of attention \cite{NIPS2017_3f5ee243} in emphasizing dynamic correlations for a long-term sequence, \cite{pan2022matt} extends a SPD attention for manifold. Specifically, we define queries from dynamic graph structures and keys and values from EEG-derived SPD matrices $\boldsymbol{K},\boldsymbol{Q},\boldsymbol{V}\in\mathbb{S}_{++}^{l}$. In the Euclidean space, the cross-attention generally defines as the the Euclidean dot-product similarity of query-key as
\[
    sim_{\text{dot}}(\boldsymbol{Q},\boldsymbol{K}) = \boldsymbol{Q}^{\text{T}}\boldsymbol{K}.
\]
However, the Euclidean dot-product can not efficiently measure the similarity because of the existing curvature in the manifold space $\mathcal{M}$. Instead, we measure the similarity of $\boldsymbol{K}$, $\boldsymbol{V} \in \mathbb{S}^{l}_{++}$ with Log-Euclidean geodesic distance on the SPD manifold, ensuring Riemannian consistency:
\[
        d_{\text{LE}}(\boldsymbol{K}_{i},\boldsymbol{Q}_{i}) =  \left\| \log(\boldsymbol{K}_{i}) - \log(\boldsymbol{Q}_{i}) \right\|_F.
\]
%The learning process is described in the Algorithm~\ref{alg:repspd}.

\subsubsection{Log-similarity Measurement}
Log-Euclidean metric offers 
\[
        d_{\text{LE}}(\boldsymbol{P}_{1},\boldsymbol{P}_{2}) =  \left\| \log(\boldsymbol{P}_{1}) - \log(\boldsymbol{P}_{2}) \right\|_F.
\]
The purpose of the $\log$ operator: $\mathbb{S}^{n}_{++} \mapsto \mathbb{S}^{n}$, that maps a SPD matrix $\boldsymbol{P} \in \mathbb{S}^{n}_{++}$ to a $n \times n$ real symmetric matrix \cite{huang2017riemannian} by:
\[
   \log({\boldsymbol{P}})= \boldsymbol{U}diag((\log(\sigma_{1}), \dots,\log(\sigma_{l}))\boldsymbol{U}^{T} ,
\]
where $\boldsymbol{U}$ is the matrix of eigenvectors of $\boldsymbol{P}$, and the eigenvalues $\sigma_{i} > 0, i = 1, ... , n$ because of $\boldsymbol{P} \in \mathbb{S}_{++}^{n}$.
We can measure the similarity of the different points \cite{NEURIPS2022_28ef7ee7} via the minimal geodesic distance:
\[
   \mathop{\arg\min}_{\boldsymbol{P} \in \mathbb{S}_{++}^{n}} \sum_{i=1}^{k} d^{2}_{\text{LE}}(\boldsymbol{P},\boldsymbol{P}_{i}) .
\]
Formally, the manifold cross-attention $\mathcal{CA}(\boldsymbol{K},\boldsymbol{Q})$ of query-key SPD matrices defines as:
\begin{equation}   
\begin{array}{rl}
   \mathcal{CA}(\boldsymbol{K},\boldsymbol{Q}) &= \mathop{\arg\min}_{\boldsymbol{K},\boldsymbol{Q} \in \mathbb{S}_{++}^{l}} \sum_{i=1}^{k} d^{2}_{\text{LE}}(\boldsymbol{K},\boldsymbol{Q}_{i}), \\[8pt]
    &= (\boldsymbol{U}_{K}\boldsymbol{U}_{Q})\|\text{Trace}\{ \log(\Sigma_{K})
    -\\[8pt]  &\quad \log(\Sigma_{Q}) \} \| (\boldsymbol{U}_{K}\boldsymbol{U}_{Q})^{\text{T}} , \\[8pt]
    &=\boldsymbol{U}_{\mathcal{CA}}(\frac{1}{L}\sum_{L}\log\|\text{det}(\Sigma_{K})-\text{det}(\Sigma)_{Q}\|) \boldsymbol{U}_{\mathcal{CA}}^{\text{T}},\\[8pt]
    &= \boldsymbol{U}_{\mathcal{CA}}(\mathbb{E}(\log(\Sigma_{K,Q}))\boldsymbol{U}_{\mathcal{CA}}^{\text{T}},
\end{array}
\end{equation}
where $\Sigma_{K} = \text{diag}(\sigma_{1}^{K}, \dots ,\sigma_{l}^{K}) $, $\Sigma_{Q} = \text{diag}(\sigma_{1}^{Q}, \dots ,\sigma_{l}^{Q}) $.

\subsubsection{Backward Procedure}
Our model integrates both Euclidean and Riemannian modules, thus requiring separate procedure in the backward propagation process \cite{huang2017riemannian}.

\textbf{In the Euclidean space}, the backward procedure follows standard chain rule-based backpropagation. Gradients of convolutional layers and GRU encoders are computed via automatic differentiation:
\[
\frac{\partial \mathcal{L}}{\partial x} = \frac{\partial \mathcal{L}}{\partial y} \cdot \frac{\partial y}{\partial x}
\]
where $x$ and $y$ are intermediate Euclidean features.

\textbf{On the Riemannian SPD manifold}, we adopt the Log-Euclidean Riemannian Metric (LERM) to ensure differentiability of SPD operations. 

\begin{enumerate}
    \item \textbf{Tangent Space Gradient:} Compute the Euclidean gradient in the log-space:
    \begin{equation}
        \nabla_{\log(\mathbf{S})} \mathcal{L} = \frac{\partial \mathcal{L}}{\partial \mathbf{V}} \cdot \frac{\partial \mathbf{V}}{\partial \log(\mathbf{S})}
    \end{equation}

    \item \textbf{Pullback to SPD Manifold:} The gradient in log-space is pulled back to the SPD manifold using the differential of the matrix logarithm:
    \begin{equation}
        \nabla_{\mathbf{S}} \mathcal{L} = \mathbf{U} \cdot \text{diag}\left( \frac{\partial \mathcal{L}}{\partial \log(\sigma)} \right) \cdot \mathbf{U}^\top
    \end{equation}
    where $\mathbf{S} = \mathbf{U} \cdot \text{diag}(\sigma) \cdot \mathbf{U}^\top$ is the eigendecomposition of $\mathbf{S}$.

    \item \textbf{Symmetrization:} Since SPD matrices lie in a symmetric space, we symmetrize the gradient to ensure it lies within the tangent space:
    \begin{equation}
        \tilde{\nabla}_{\mathbf{S}} \mathcal{L} = \frac{1}{2} \left( \nabla_{\mathbf{S}} \mathcal{L} + \nabla_{\mathbf{S}} \mathcal{L}^\top \right)
    \end{equation}
\end{enumerate}
This gradient formulation respects the manifold geometry and curvature, ensuring stable and meaningful gradient updates over SPD representations.

\section{Implementation and Model Training}
\subsubsection{Hyperparameter Settings}
We implement our model with PyTorch and Pytorch Geometric libraries platform and conduct all experiments on an NVIDIA A6000 GPU. For EEG encoding, we use a spaotemporal CNN backbone with kernel size of (19,1) and 64 hidden units with temporal dimension pooling, and (1,31) and project 64 to 20 outchannels , followed by a GRU layer with 64 hidden states to capture temporal dependencies across the EEG segments. The SPD matrices are constructed from the output of the encoder with a regularization coefficient $\epsilon = 1e$-$4$ to ensure numerical stability. For the SPDNet backbone, we use two BiMap layers with dimensions (20, 8), followed by a LogEig layer. The cross-attention module computes query-key interactions using log-Euclidean distance and outputs modulated SPD representations with dimension $8 \times 8 $. During training, we set the batch size to 128, and optimize the model using Adam optimizer with a learning rate of 0.001. The total number of epochs is set to 150, and the geometric consistency loss $\mathcal{L}_{\text{GeoTop}}$ is weighted with $\beta = 0.3$.

\subsection{Datasets}
Dataset Seizure: We evaluate our method on the Temple University Hospital EEG Seizure Corpus (TUSZ v1.5.2) \cite{shah2018temple}, one of the largest publicly available clinical EEG datasets for seizure detection \cite{shah2018temple}. TUSZ contains over 5,600 EEG recordings with 3,050 manually annotated seizures, covering a wide spectrum of seizure types and clinical scenarios. Each recording consists of $19$ EEG channels following the $10$-$20$ system, ensuring consistency with clinical standards.
To simulate a realistic seizure detection, we segment each EEG recording into $1$s time-window length with 12 EEG segments. For early prediction tasks, we define the one-minute period prior to seizure onset as the preictal segment, representing the critical phase for early intervention. A five-minute buffer zone around the boundary of seizure onset is excluded from the training to avoid ambiguous data leakage.

%This dataset presents significant challenges due to its temporal heterogeneity, channel variability, and lack of preictal ground-truth labels, making it a suitable benchmark for evaluating the robustness and generalization capability of brain dynamics models.

Dataset MI: We utilize the BNCI2014001 dataset from the MOABB benchmark \cite{Aristimunha_Mother_of_all}, which provides a standardized and extensively used EEG motor imagery corpus. This dataset includes EEG recordings from multiple subjects performing a binary motor imagery task (left hand vs. right hand). We adopt the Motor Imagery paradigm with two classes: right hand and left hand, and a frequency band of 8–30 Hz. EEG signals are sampled from 22 channels at a native frequency of 200 Hz and resampled to 250 Hz for consistency.
During each trial, the subject was instructed to imagine the movement of four classes from 0.5 to 4.0 seconds following a visual cue. Preprocessing steps include: 
\begin{itemize}
    \item resampling to 200 Hz,
    \item  band-pass filtering between 8–30 Hz,
    \item  epoch segmentation from 0.5s to 4.0s.
\end{itemize}
After these steps, the format of the training EEG data is in 5 segments with 200 time points (i.e., $0.4$ s/epoch) per trial. The extracted EEG segments form the input for our SPD and graph-based modeling pipeline.

\section{Additional results}
We conduct an ablation study to examine the effect of different projection mechanisms and geometric losses in our SPD-based attention framework. As shown in Table~\ref{tab:ablation_comparison_detection_12s}, our proposed method, \method, outperforms all ablation variants with an accuracy of 86.4\%, AUROC of 81.6, and F1-score of 42.9, demonstrating strong robustness and generalization. Notably, using QR and SVD-based projectors alone underperforms compared to the full method. For example, QR-projector and SVD-projector achieve only 83.5\% and 79.6\% accuracy, respectively. Moreover, removing the geometric alignment loss $\mathcal{L}_{\text{GeoTop}}$ leads to significant performance drops (e.g., F1 from 35.5\% → 33.5\% in QR case), confirming its importance in preserving topological structure on the manifold. Furthermore, when replacing manifold attention with a Euclidean cross-attention baseline, performance also drops to 38.4\%, indicating that geometry-aware attention better captures intrinsic relationships in brain dynamics.
\begin{table}[t]
\centering
%\begin{threeparttable}
\caption{Ablation comparison on TUSZ dataset for 12s seizure detection with QR, SVD-projector, and regular cross-attention. \textbf{Bold} and \underline{underline} indicate best best results.}
\label{tab:ablation_comparison_detection_12s}
  {\fontsize{8pt}{11pt}\selectfont
   \resizebox{0.5\linewidth}{!}{%
\begin{NiceTabular}{llccc} 
       \toprule[1.2pt]
       \rowcolor{gray!20}
Method & Type          &ACC  & AUROC   & F1    \\
       \midrule
QR-projector          & Manifold            & 83.5    & 76.3 & 35.5 \\
QR-projector w/o $\mathcal{L}_{\text{GeoTop}}$          & Manifold            & 79.0    & 74.7 & 33.5 \\

SVD-projector        & Manifold          & 79.6     & 77.2  & 34.1  \\
SVD-projector w/o $\mathcal{L}_{\text{GeoTop}}$        & Manifold          & 72.6     & 73.8  & 30.3  \\
cross-attention     & Euclidean  & 84.0& 79.6 & 38.4 \\
%SPDNet           & Manifold     &84.1 & 78.2 & 30.2 \\
%mAtt          & Manifold     & \underline{84.7} & 80.5 & 38.6  \\
\midrule
\method & Manifold &\textbf{86.4} ($\uparrow2.4$) & \textbf{81.6} ($\uparrow2.0$) & \textbf{42.9} ($\uparrow4.5$)  \\ 

%\textbf{\method (Ours-Temporal+Graph)} &  & \textbf{0.833} & \textbf{0.434}& \textbf{0.350} \\
\bottomrule
\end{NiceTabular}
   }% end resizebox
  }% end fontsize

\end{table}

\begin{figure*}[htbp]
\centering
\includegraphics[width = 0.85\textwidth]{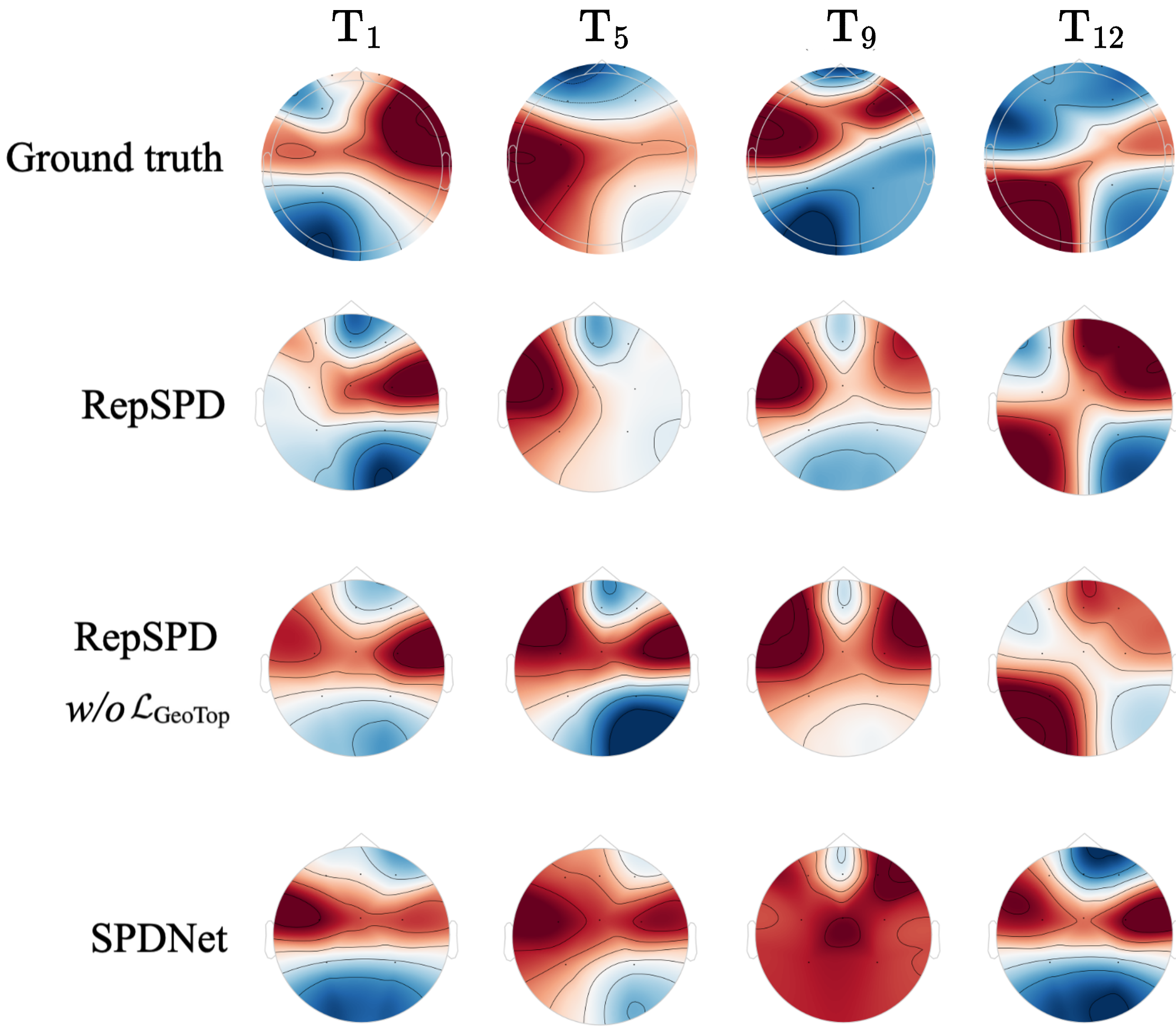}
\caption{Topographic visualizations of spatial brain activation at selected time segments ($T_1$, $T_5$, $T_9$, $T_{12}$) from one EEG trial. Each map illustrates the activation over 19 channels projected to 2D scalp topography.
We compare the output of our method (\method), the ablated variant without geometric alignment loss ($\mathcal{L}_{\text{GeoTop}}$), and the baseline SPDNet against the ground truth.}
\label{fig: app_brain}
\end{figure*}

\begin{table}[htbp] 
\centering
\caption{Ablation comparison with meta optimization.}
{\fontsize{8pt}{11pt}\selectfont
   \resizebox{0.5\linewidth}{!}{%
\begin{NiceTabular}
    {l c c c c} 
    \toprule[1.2pt]
    \rowcolor{gray!20}
Method & Mode & ACC & AUROC & F1 \\
       \midrule
\method w & Train & 89.4 & 81.6 & 45.2 \\
\method w & Test & 86.4 & 81.6 & 42.9 \\
\method w/o &Train& 88.4 & 83.0 & 43.4 \\
\method w/o & Test & 78.8($\downarrow9.6$) & 73.6 ($\downarrow9.4$) & 34.8 ($\downarrow8.6$)\\
\bottomrule
\end{NiceTabular}
}
}
\label{tab: opt abl}
\end{table}

To investigate the effectiveness of our meta-optimization strategy that combines Riemannian and Euclidean gradients, we conduct an ablation analysis as shown in Table~\ref{tab: opt abl}.
We compare the performance of \method with and without the meta-optimization during both training and testing phases.
\method w/o meta-optimization exhibits severe performance degradation on the test set in ACC ($7.8\%$) and F1 ($8.6 \%$), despite achieving high accuracy on the training set at $88.4 \%$. This highlights the risk of overfitting when optimization is confined to either the manifold or Euclidean space alone.
In contrast, \method with meta-optimization maintains consistent performance between training and test phases, suggesting that alternating between Riemannian (for SPD representation) and Euclidean (for neural weights) optimization improves generalization.
These findings demonstrate that the hybrid strategy facilitates stable gradient propagation and effectively bridges the geometric gap between SPD-based structures and neural modules.

Figure~\ref{fig: app_brain} further analyzes the effect of graph-guided SPD modulation, we visualize the evolution of spatial brain patterns over time.
We select one EEG sample from the TUSZ dataset with 12 segments.
We sample four representative frames ($T_1$, $T_5$, $T_9$, $T_{12}$) with a stride of 3, and visualize their corresponding topographic brain maps across models.
Our proposed \method exhibits significantly improved alignment with ground-truth topologies, preserving dynamic spatial structures throughout the timeline.
In contrast, SPDNet fails to adaptively track local connectivity variations, showing oversmoothed activations.
Moreover, removing geometric alignment loss $\mathcal{L}_{\text{GeoTop}}$ results in blurry and unstable patterns, confirming its importance in the regulation of curvature-induced distortions. This visualization demonstrates that integrating dynamic topology into SPD representation via manifold cross-attention enables a more expressive and temporally aware modeling of functional brain connectivity.

\begin{table*}[htbp]
\centering
\caption{Ablation on TUSZ dataset for 12s seizure detection with different top-$\tau$ options. \textbf{Bold} and \underline{underline} indicate best and second-best results.}
\label{tab:ab_top_tau}
\begin{tabular}{ccccc}  
\toprule
Top-$\tau$           & AUROC   & ACC      & F1  \\
\midrule
 2                 & 0.807 $\pm$ 0.003   & 0.815$\pm$0.003  & 0.404$\pm$0.009 \\
3                &  \textbf{0.816$\pm$0.006}  & \textbf{0.864$\pm$0.003}  & \textbf{0.429$\pm$0.017}  \\
7    & \underline{0.809$\pm$0.004}  & \underline{0.862$\pm$0.004}  & 0.411$\pm$0.013  \\
9          & 0.801$\pm$0.004  & 0.849$\pm$0.004  & 0.417$\pm$0.011  \\
11 & 0.776$\pm$0.004 & 0.821$\pm$0.002 & \underline{0.419$\pm$0.003} \\ 
13 & 0.805$\pm$0.003 & 0.832$\pm$0.004 & 0.414$\pm$0.003 \\

%\textbf{\method (Ours-Temporal+Graph)} &  & \textbf{0.833} & \textbf{0.434}& \textbf{0.350} \\
\bottomrule
\end{tabular}
\end{table*}

Table~\ref{tab:ab_top_tau} studies the effect of the top-$\tau$ option in our method on TUSZ seizure detection. The top-$\tau$ controls the strongest connections at each epoch for sparse and meaningful graph representations to avoid redundant connections and clear spatial structures. Overall, a moderate top-$\tau$ yields the best performance, while overly small or large values lead to degraded results. In particular, top-$\tau$=3 achieves the best AUROC of 0.816±0.006 and the best F1 of 0.429±0.017. When $\tau$ becomes larger, both AUROC and F1 decrease, suggesting that selecting too many candidates may dilute discriminative patterns and introduce noise. These results indicate that top-$\tau$ controls an important trade-off between capturing informative structures and avoiding over-inclusive selections, and we adopt top-$\tau$=3 as the default setting.


% xxxx

%%%%%%%%%%%%%%%%%%%%%%%%%%%%%%%%%%%%%%%%%%%%%%%%%%%%%%%%%%%%%%%%%%%%%%%%%%%%%%%
%%%%%%%%%%%%%%%%%%%%%%%%%%%%%%%%%%%%%%%%%%%%%%%%%%%%%%%%%%%%%%%%%%%%%%%%%%%%%%%

\end{document}